\documentclass[letterpaper, 10 pt, conference]{ieeeconf}  % Comment this line out if you need a4paper

\IEEEoverridecommandlockouts                              % Needed for the \thanks command
\usepackage{colortbl}
\usepackage{color}
\usepackage{tabularray}
\usepackage{subcaption}
\usepackage{multirow}
\let\labelindent\relax
\usepackage{enumitem}
\usepackage{graphicx} 
\usepackage{balance}
\usepackage[table,xcdraw]{xcolor}
\usepackage[noadjust]{cite}
\usepackage{hyperref}
\usepackage{svg}

\title{\LARGE \bf Toward RAPS: the Robot Autonomy Perception Scale}

\author{Rafael Sousa Silva, Cailyn Smith, Lara Bezerra, and Tom Williams%
\thanks{MIRRORLab, Colorado School of Mines, Golden, CO 80401, USA.
        {\tt\small rsousasilva@mines.edu}, 
        {\tt\small ccsmith1@mines.edu},
        {\tt\small lbezerra@mines.edu},
        {\tt\small twilliams@mines.edu}}%
}

\begin{document}

\maketitle
\thispagestyle{empty}
\pagestyle{empty}

%%%%%%%%%%%%%%%%%%%%%%%%%%%%%%%%%%%%%%%%%%%%%%%%%%%%%%%%%%%%%%%%%%%%%%%%%%%%%%%%
\begin{abstract}
Human-robot interactions can change significantly depending on how autonomous humans perceive a robot to be. Yet, while previous work in the HRI community measured perceptions of \textit{human} autonomy, there is little work on measuring perceptions of \textit{robot} autonomy. In this paper, we present our progress toward the creation of the Robot Autonomy Perception Scale (RAPS): a theoretically motivated scale for measuring human perceptions of robot autonomy. We formulated a set of fifteen Likert scale items that are based on the definition of autonomy from Beer et al.'s work, which identifies five key autonomy components: ability to sense, ability to plan, ability to act, ability to act with an intent towards some goal, and an ability to do so without external control. We applied RAPS to an experimental context in which a robot communicated with a human teammate through different levels of Performative Autonomy (PA): an autonomy-driven strategy in which robots may ``perform" a lower level of autonomy than they are truly capable of to increase human situational awareness. Our results present preliminary validation for RAPS by demonstrating its sensitivity to PA and motivate the further validation of RAPS.
\end{abstract}

%%%%%%%%%%%%%%%%%%%%%%%%%%%%%%%%%%%%%%%%%%%%%%%%%%%%%%%%%%%%%%%%%%%%%%%%%%%%%%%%

\section{INTRODUCTION AND BACKGROUND}

When humans interact with a robot, they form a mental model of how autonomous they believe the robot is and create expectations regarding its capabilities~\cite{nass2000machines}.
For example, if they see that a robot is equipped with a camera, it would be reasonable to assume that it may be able to look at someone's face or see an object that is placed in front of it.
Yet, these mental models often underestimate or overestimate the robot's true capabilities, leading to suboptimal use of the robot or human disappointment~\cite{harbers2017perceived}.
As such, the way in which humans perceive robot autonomy can shape different aspects of interaction.
While some work in the HRI community measures perceptions of \textit{human} autonomy~\cite{sankaran2021exploring,wilson2023perspectives}, as Kim et al.~\cite{kim2023perspectives} point out, there is little work on measuring perceptions of \textit{robot} autonomy, despite it being a key factor in determining the \textit{agency} of a robot at a particular level of abstraction~\cite{floridi2004morality,jackson2021theory}.
The most relevant prior work on measuring perceived \textit{robot} autonomy is that of Harbers et al.~\cite{harbers2017perceived}, who analyze autonomy with specific focus on the temporal extent of robot independence, the extent to which robots are obedient, and the extent to which robots inform users of their goals and plans. 
However, these authors explore the factors that \textit{impact} perceived autonomy, with perceived autonomy itself only being measured through a single item: ``How autonomous do you consider this system?''. 
Given the problems with single-item measures~\cite{schrum2020four}, especially for multi-dimensional HRI concepts~\cite{malle2021multidimensional}, this suggests a significant need for a multi-dimensional measure of perceived autonomy for HRI.
Understanding how humans perceive robot autonomy is critical, as robots with \textit{actually} lower levels of autonomy may be perceived as less intelligent~\cite{choi2014autonomy}, and lower levels of autonomy and intelligence (a key dimension of robots' expert power~\cite{hou2024power}) may lead to lower levels of human-robot trust~\cite{desai2012effects}.

The development of a scale is an iterative process composed of multiple steps that range from the creation and evaluation of scale items to final assessments of scale validity and reliability~\cite{kyriazos2018applied, boateng2018best}.
In this work, we thus present the initial steps toward the creation of the Robot Autonomy Perception Scale (RAPS), a theoretically motivated scale for measuring perceived robot autonomy.
We provide motivation for creating adequate scale items and assess whether the scale detects changes in perceived autonomy when a robot adopts an autonomy manipulation strategy from literature.
Given the dearth of tools for measuring perceived autonomy, with the development of RAPS we ultimately seek to answer:
\begin{enumerate}[label=\textbf{RQ:}]
    \item How can perceived robot autonomy be measured through a multi-dimensional psychometric scale? 
\end{enumerate}

\section{RAPS SCALE DEVELOPMENT}
To answer our research question, we began by considering the wide variety of available definitions for autonomy (see~\cite{kim2024taxonomy} for an extensive discussion on how this concept has been defined and used in literature). 
Within this space of definitions, we selected that proposed by Beer et al.~\cite{beer2014toward} to serve as inspiration for the overall structure of our scale: ``The extent to which a robot can \textbf{sense} the environment, \textbf{plan} based on that environment, and \textbf{act} upon that environment, with the intent of reaching some \textbf{goal} (either given to or created by the robot) without external \textbf{control}." 
We found that this definition was the most complete for two reasons.
First, it aggregates five key dimensions of robot autonomy: ability to sense, ability to plan, ability to act, ability to act with an intent towards some goal, and an ability to do so without external control.
Second, it brings together four of the most important \textit{forms} of autonomy within Human-Robot Interaction, as per Kim et al.~\cite{kim2024taxonomy}. 
The ability to \textit{sense and plan} represent different dimensions of \textit{cognitive autonomy}. 
The ability to \textit{act} corresponds with \textit{physical autonomy}. 
The ability to act in a \textit{goal directed fashion} corresponds with \textit{intentional autonomy}. 
And the ability to act in a way that is self-determined, without external \textit{control}, corresponds to \textit{operational autonomy}. 
We argue that a multi-component scale grounded in these five elements would be an effective way of jointly measuring -- yet also teasing apart -- these key forms of robot autonomy.

As such, we next moved to identify scale items that could be used to measure each of these five components. 
For each component, we designed scale items around definitions of autonomy found across literature. 
From among these items, we down-selected to include the three most relevant items for each of the scale components, producing a complete set of fifteen scale items. 
The final 15-item Robot Autonomy Perception Scale (RAPS) is shown in Table~\ref{tab:scale}, along with the citations for the papers that inspired each item.
Each scale item is intended to be presented as a 7-point Likert item (1=Strongly Disagree to 7=Strongly Agree).
With the creation of this final set of measures, we applied RAPS to an experimental context to assess the scale's sensitivity to an autonomy manipulation strategy from literature.

\begin{table}[t]
    \centering
    \begin{tabular}{c|p{5cm}|c}
        \textbf{Component} & \textbf{Item} & \textbf{Source} \\\hline
        \multirow{3}{*}{Sense} & The robot is able to sense its surroundings. & \cite{franklin1996agent} \\\cline{2-3} 
 & The robot is able to detect variations in its surroundings. & \cite{thrun2004toward} \\\cline{2-3} 
 & The robot is able to perceive its current situation. & \cite{huang2004autonomy} \\\hline 
        \multirow{3}{*}{Plan} & The robot is able to establish a course of action. & \cite{alami2005task} \\\cline{2-3} 
 & The robot is able to plan its behavior. & \cite{huang2004autonomy} \\\cline{2-3} 
 & The robot is able to plan according to changes in its surroundings. & \cite{murphy2019introduction} \\\hline
        \multirow{3}{*}{Act} & The robot is able to carry out its own actions. & \cite{alami1998architecture} \\\cline{2-3} 
 & The robot is able to take actions to affect its environment. & \cite{beer2014toward} \\\cline{2-3} 
 & The robot is able to take actions that allow others to understand the current situation. & \cite{luebbers2023autonomous} \\\hline
        \multirow{3}{*}{Goal} & The robot has the intent to achieve goals. & \cite{beer2014toward} \\\cline{2-3} 
 & The robot is capable of setting goals for itself. & \cite{huang2004autonomy} \\\cline{2-3} 
 & The robot would modify its behavior if this were necessary to pursue its goals. & \cite{murphy2019introduction} \\\hline
        \multirow{3}{*}{Control} & The robot makes its own choices. & \cite{formosa2021robot} \\\cline{2-3} 
 & The robot can operate without the direct intervention of humans. & \cite{wooldridge1995intelligent} \\\cline{2-3} 
 & The robot exercises control over its own actions. & \cite{franklin1996agent} \\\hline
    \end{tabular}
    \caption{Robot Autonomy Perception Scale (RAPS)}
    \label{tab:scale}
\end{table}

%%%%%%%%%%%%%%%%%%%%%%%%%%%%%%%%%%%%%%%%%%%%%%%%%%%%%%%%%%%%%%%%%%%%%%%%%%%%%%%%
\section{PRELIMINARY VALIDATION} \label{sec:experiment}

To obtain preliminary validation, we applied RAPS to an experimental context (N=76) in which a robot communicated with a human teammate through different levels of Performative Autonomy (PA)\footnote{More information about the experiment and its results will be submitted as a full paper to an upcoming conference.}.
PA was introduced by Roy et al.~\cite{roy2023need} as a robotic strategy for increasing human teammates' situational awareness in safety-critical domains without increasing their cognitive load. In their experiments, they found supporting evidence that the use of PA indeed benefits human situational awareness.
In collaborative scenarios where human loss of awareness can be costly (e.g., search and rescue), robots using PA may ``perform" a lower level of autonomy than they are truly capable of.
They do so by asking their human teammates questions that they do not believe they need the answers to (using lower levels of dialogue autonomy, as shown in Table~\ref{tab:dialogueAutonomy}), but that highlight important aspects of the robot's task.

\begin{table}[t]
    \centering
    \begin{tabular}{c|p{5cm}|p{1.8cm}}
         \textbf{Level} & \textbf{Strategy} & \textbf{Speech Act} \\\hline
         6 & Selecting option without proposal & (None)\\ \hline
         5 & Proposing and selecting a single option without opportunity for veto & Requests/ Commands\\ \hline
         4 & Proposing and selecting a single option with opportunity for veto & Statements/ Assertions\\ \hline
         3 & Proposing a single option & Suggestions\\ \hline
         2 & Requesting confirmation of a single option &YN-Questions\\ \hline
         1 & Requesting selection between multiple options & WH-Questions\\ \hline
    \end{tabular}
    \caption{Dialogue Autonomy Levels \& Associated Speech Acts (From~\cite{roy2023need})}
    \label{tab:dialogueAutonomy}
\end{table}

Given the applications of PA, in previous work Sousa Silva et al.~\cite{sousasilva2023worth} have assessed its impact on team performance, cognitive load, and perceptions of robot teaming quality and dependency in collaborative tasks.
Because their work was motivated by space exploration scenarios, human-robot communication in their experiment was hindered by latency to account for the delays that happen when messages are sent and received in such scenarios.
While their work showed that PA strategies can be beneficial for teaming quality and cognitive load stability in collaborative scenarios that suffer from latency, it did not demonstrate whether \textit{performing} lower levels of autonomy led users to actually \textit{perceive} lower levels of robot autonomy. 
As such, in our first experiment for the preliminary validation of RAPS, we applied our scale to the same experimental context used in~\cite{sousasilva2023worth} to assess whether human perceptions of autonomy are actually affected by PA.
In addition, we were interested in observing through our results which sub-components of RAPS would be sensitive to PA given the chosen experimental design.

\subsection{Experimental Design}

\begin{figure}[h]
    \centering
    \includegraphics[width=0.75\linewidth]{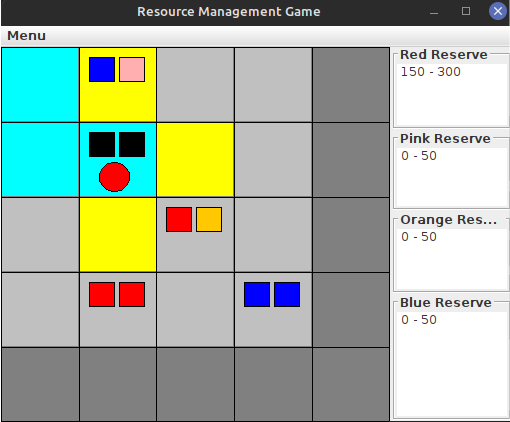}
    \caption{Reduced visualization of the player game interface. Resource stations show the type of resources needed to clear that spot. Cleared resource stations turn black.}
    \label{fig:game}
\end{figure}

Participants were tasked with completing the collaborative resource management game previously used in \cite{sousasilva2023worth}.
In this task, participants play a game where they spend four different types of resources that are collected by a remote robot teammate (a Misty II). 
Because of the robot's remote location, participants experienced latency when sending and receiving messages from the robot.
They could determine the rough levels of each resource available through indicators shown on their screen (see Figure~\ref{fig:game}).
At any time, participants could identify which resource the robot was collecting by observing an additional monitor with a live stream of the robot.
Periodically, the robot would re-assess which resource was most strategically needed, and then either inform the participant that it was switching to collect that resource, or ask the participant as to which resource it should collect, depending on experimental condition, as explained below.
Participants were responsible for spending resources in stations spread throughout the board and the game finished after all available stations were cleared.

Our experiment followed a Greco-Latin Square design in which \textit{Latency} varied on three levels (\textit{Low Latency} [0s], \textit{Medium Latency} [5s], and \textit{High Latency} [10s]) and \textit{PA Strategy} varied on three levels (\textit{Low PA} [level 1 of dialogue autonomy], \textit{Medium PA} [level 2 of dialogue autonomy], and \textit{High PA} [level 5 of dialogue autonomy]).
For each PA strategy, the robot communicated with participants about resource collection in the following manner:
\begin{itemize}
    \item \textit{Low PA}: the robot asked for arbitration regarding which resource it should collect next, given three out of four possibilities (e.g., ``I was collecting orange resources. Should I keep collecting orange resources, or switch to pink resources, or to red resources?"). 
    \item \textit{Medium PA}: the robot selected a follow up action and asked participants for confirmation through a Yes/No question (e.g., ``I was collecting red resources. Can I now switch to collecting blue resources?").
    \item \textit{High PA}: the robot selected a follow up action and stated it to the participants without receiving any input (e.g., ``I was collecting pink resources. I am now going to collect orange resources.").
\end{itemize}

As in \cite{sousasilva2023worth}, messages were conveyed to participants through both text and audio.
When robot-to-human communication was performed under latency, a time delay was imposed to simulate the robot's communication traveling to the human teammate; 
when a human-to-robot response was provided, an additional time delay was imposed to represent the time needed for the message to reach the robot and for the audiovisual data of the robot to be ostensibly relayed back.
Each participant played three games, experiencing all possible levels of latency and PA strategies.
After each game, participants were asked to fill out a survey that included RAPS to indicate their perceptions of the robot's autonomy.

%%%%%%%%%%%%%%%%%%%%%%%%%%%%%%%%%%%%%%%%%%%%%%%%%%%%%%%%%%%%%%%%%%%%%%%%%%%%%%%%
\section{RESULTS} \label{sec:results}

\subsection{Reliability Analysis}
After collecting all datapoints, we measured Cronbach's alpha to assess the internal consistency of RAPS.
The result, $\alpha=0.928$, supports the reliability of RAPS for measuring perceived autonomy. 

\subsection{Bayesian Statistical Analysis}
\begin{figure}[t]
    \centering
    \includegraphics[width=\linewidth]{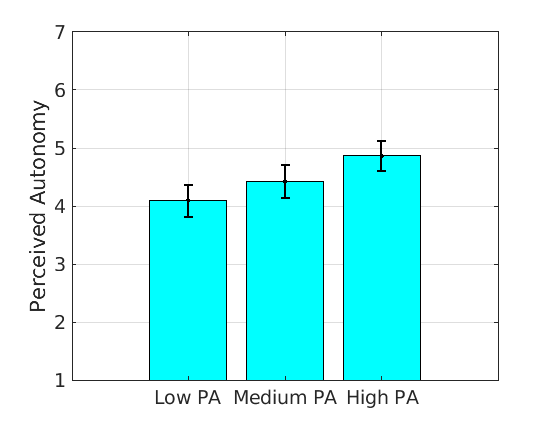}
    \caption{Effects of PA Strategy on perceived autonomy. Error bars represent 95\% confidence interval.}
    \label{fig:autonomy}
\end{figure}

Table~\ref{tab:descriptives} displays the mean and standard deviation values for participants in each experimental condition.
To determine whether our autonomy manipulations led to measurable differences in RAPS scores, we conducted Repeated Measures (RM) Bayesian ANOVAs with inclusion Bayes Factor ($BF_{10}$) analysis. 
These analyses revealed extreme evidence for an impact of PA strategy on perceived autonomy as measured by the RAPS scale ($BF_{10} = 5.34\times10^5$), anecdotal evidence against an effect of Latency ($BF_{10} = 0.744$), and moderate evidence against an interaction effect ($BF_{10} = 0.250$).
Post hoc tests (see Table~\ref{tab:posthocs}) revealed that higher PA lead to higher perceived autonomy ratings.
The descriptive statistics for PA strategy results are visualized in Figure~\ref{fig:autonomy} and are based on perceived autonomy scores calculated by averaging all RAPS scale items.

\begin{table}[t]
\centering
\resizebox{0.75\columnwidth}{!}{%
\begin{tabular}{|c|cc|}
\hline
 & \multicolumn{2}{c|}{\cellcolor[HTML]{C0C0C0}\textbf{Perceived Autonomy}} \\ \cline{2-3} 
\multirow{-2}{*}{} & \multicolumn{1}{c|}{\cellcolor[HTML]{EFEFEF}Mean} & \cellcolor[HTML]{EFEFEF}SD \\ \hline
\cellcolor[HTML]{EFEFEF}\textbf{Low Latency (LL)} & \multicolumn{1}{c|}{4.513} & 1.191 \\ \hline
\cellcolor[HTML]{EFEFEF}\textbf{Medium Latency (ML)} & \multicolumn{1}{c|}{4.584} & 1.170 \\ \hline
\cellcolor[HTML]{EFEFEF}\textbf{High Latency (HL)} & \multicolumn{1}{c|}{4.291} & 1.292 \\ \hline\hline
\cellcolor[HTML]{EFEFEF}\textbf{Low PA (LPA)} & \multicolumn{1}{c|}{4.100} & 1.241 \\ \hline
\cellcolor[HTML]{EFEFEF}\textbf{Medium PA (MPA)} & \multicolumn{1}{c|}{4.425} & 1.254 \\ \hline
\cellcolor[HTML]{EFEFEF}\textbf{High PA (HPA)} & \multicolumn{1}{c|}{4.864} & 1.157 \\ \hline\hline
\cellcolor[HTML]{EFEFEF}\textbf{LL + LPA} & \multicolumn{1}{c|}{3.831} & 1.335 \\ \hline
\cellcolor[HTML]{EFEFEF}\textbf{LL + MPA} & \multicolumn{1}{c|}{4.944} & 1.182 \\ \hline
\cellcolor[HTML]{EFEFEF}\textbf{LL + HPA} & \multicolumn{1}{c|}{4.765} & 1.055 \\ \hline
\cellcolor[HTML]{EFEFEF}\textbf{ML + LPA} & \multicolumn{1}{c|}{4.165} & 1.306 \\ \hline
\cellcolor[HTML]{EFEFEF}\textbf{ML + MPA} & \multicolumn{1}{c|}{4.241} & 1.262 \\ \hline
\cellcolor[HTML]{EFEFEF}\textbf{ML + HPA} & \multicolumn{1}{c|}{5.347} & 0.941 \\ \hline
\cellcolor[HTML]{EFEFEF}\textbf{HL + LPA} & \multicolumn{1}{c|}{4.304} & 1.083 \\ \hline
\cellcolor[HTML]{EFEFEF}\textbf{HL + MPA} & \multicolumn{1}{c|}{4.091} & 1.318 \\ \hline
\cellcolor[HTML]{EFEFEF}\textbf{HL + HPA} & \multicolumn{1}{c|}{4.479} & 1.476 \\ \hline
\end{tabular}%
}
\caption{Mean and Standard Deviation values for each analysis group}
\label{tab:descriptives}
\end{table}

% Latency and PA Strategy Posthocs
\begin{table*}[h]
\centering
\begin{tabular}{|ccc||ccc|}
\hline
\rowcolor[HTML]{C0C0C0} 
\multicolumn{3}{|c||}{\cellcolor[HTML]{C0C0C0}\textbf{Latency}} & \multicolumn{3}{c|}{\cellcolor[HTML]{C0C0C0}\textbf{PA Strategy}} \\ \hline
\rowcolor[HTML]{EFEFEF} 
\multicolumn{1}{|c|}{\cellcolor[HTML]{EFEFEF}\textbf{\begin{tabular}[c]{@{}c@{}}Low Latency (LL)\\ v.\\ Medium Latency (ML)\end{tabular}}} & \multicolumn{1}{c|}{\cellcolor[HTML]{EFEFEF}\textbf{\begin{tabular}[c]{@{}c@{}}LL\\ v.\\ High Latency (HL)\end{tabular}}} & \textbf{\begin{tabular}[c]{@{}c@{}}ML\\ v.\\ HL\end{tabular}} & \multicolumn{1}{c|}{\cellcolor[HTML]{EFEFEF}\textbf{\begin{tabular}[c]{@{}c@{}}Low PA (LPA)\\ v.\\ Medium PA (MPA)\end{tabular}}} & \multicolumn{1}{c|}{\cellcolor[HTML]{EFEFEF}\textbf{\begin{tabular}[c]{@{}c@{}}LPA\\ v.\\ High PA (HPA)\end{tabular}}} & \textbf{\begin{tabular}[c]{@{}c@{}}MPA\\ v.\\ HPA\end{tabular}} \\ \hline
\multicolumn{1}{|c|}{{\color[HTML]{C0C0C0} 0.201}} & \multicolumn{1}{c|}{0.411} & 1.84 & \multicolumn{1}{c|}{3.10} & \multicolumn{1}{c|}{\textbf{2.86e+05}} & \textbf{75.03} \\ \hline
\end{tabular}
\caption{Inclusion Bayes Factors ($BF_{10}$) for Latency and PA Strategy. Results with conclusively positive evidence are bolded; Results with conclusively negative evidence are grayed out.}
\label{tab:posthocs}
\end{table*}

To develop a deeper understanding of these results, we further analyzed the effects of PA on each of the sub-components of RAPS (i.e., by only averaging the scores provided to the three items on a scale sub-component). 
In doing so, we observed strong evidence for effects of PA on participants' ratings for the robot's ability to plan ($BF_{10} = 10.76$), very strong evidence for effects of PA on ratings for the robot's ability to act ($BF_{10} = 88.79$), and extreme evidence for effects of PA on ratings for the robot's ability to operate without external control ($BF_{10} = 1.1\times10^7$). 
Specifically, at an item-level analysis, we identified that when lower PA strategies were used participants thought the robot was less able to (1) establish a course of action, 
(2) carry out its own actions, 
(3) make its own choices, 
(4) operate without direct human intervention, and 
(5) exercise control over its own actions.
No effects were found on the robot's ability to sense or the robot's goal-directedness.

%%%%%%%%%%%%%%%%%%%%%%%%%%%%%%%%%%%%%%%%%%%%%%%%%%%%%%%%%%%%%%%%%%%%%%%%%%%%%%%%
\section{DISCUSSION} \label{sec:discussion}

In this work, we provided initial motivation for the creation of a new scale for perceived robot autonomy and designed an initial set of fifteen scale items that are based on definitions of autonomy from literature.
We confirmed that robots were indeed perceived to be less autonomous in ways that were measurable by RAPS when they were performing lower PA strategies, although the observable differences were not terribly large (4.1 vs 4.4 vs 4.8 / 7).
This finding reinforces \cite{roy2023need}'s assertion that different levels of autonomy can be performed (and thus perceived) through different types of dialogue moves.

Furthermore, our component-level analysis of RAPS scores suggests other key insights both about PA, and about RAPS.
Notably, our results suggest that PA specifically forms three of the four included key types of autonomy from Kim et al.'s taxonomy: cognitive autonomy, physical autonomy, and operational autonomy (excluding intentional autonomy)~\cite{kim2024taxonomy}.
In addition, results indicated that PA performs three of the five key dimensions of autonomy from Beer et al.'s definition~\cite{beer2014toward}, while not necessarily performing the other two dimensions. 
Participants' responses suggest that the robot's abilities to plan, to act, and to operate without external control were important for this specific experimental context.
These results were expected, as the adoption of lower PA strategies provided participants with
more autonomy to plan the robot's collection strategy, to direct the robot's actions, and to
control the robot's decisions.

Furthermore, participants' ratings for the robot's ability to sense and its goal-directedness did not seem to change across conditions in our experimental context.
On one hand, we expected the adoption of lower PA strategies to have no effects on how the robot was sensing the environment. 
After all, this ability was indeed not directly conveyed through PA strategy changes nor through robot communication.
On the other hand, we \textit{did} expect to see differences in participants' ratings for goal-directedness, as robots using lower PA strategies could be interpreted as less capable of setting their own goals without human input.
It is possible that the specific items used by RAPS for goal-directedness were not sufficiently diagnostic. 
This suggests a need for future work to re-evaluate certain scale items and to explore experimental contexts that are different from those examined in this work.

Finally, our findings motivate future work to build on these initial steps to complete the creation process for scale items.
After completing this step, future work should seek further validation by considering the assessments of experts and study participants, performing tests of dimensionality, validity, and reliability of the RAPS scale items, and following other steps required to develop a multi-dimensional scale that have been suggested by literature~\cite{boateng2018best, kyriazos2018applied}.

%%%%%%%%%%%%%%%%%%%%%%%%%%%%%%%%%%%%%%%%%%%%%%%%%%%%%%%%%%%%%%%%%%%%%%%%%%%%%%%%
\section{CONCLUSION} \label{sec:conclusion}
In this study, we introduced and obtained preliminary validation for the Robot Autonomy Perception Scale (RAPS), a multi-component scale based on the definition of autonomy put forward by Beer et al.~\cite{beer2014toward}, which captures four of the six dimensions of autonomy identified by Kim et al.~\cite{kim2024taxonomy}.
We used RAPS on a human-subjects experiment to assess participants' perceptions of robot autonomy when different Performative Autonomy strategies were in use.
Our results suggest that RAPS is indeed sensitive to these Performative Autonomy strategies, which may be adopted by robots to increase human situational awareness and preserve their cognitive load.
Specifically, RAPS accurately detected changes in participants' perceptions of robot autonomy with regards to the robot's abilities to plan, act, and operate without external control.
Yet, the validation of the RAPS scale in this work is preliminary, and only the first step towards full validation.
As such, future work is encouraged to perform further scale validation steps that have been suggested by previous authors, such as including experts and study participants into the design process and performing further item reliability analysis.

%%%%%%%%%%%%%%%%%%%%%%%%%%%%%%%%%%%%%%%%%%%%%%%%%%%%%%%%%%%%%%%%%%%%%%%%%%%%%%%%
\section*{ACKNOWLEDGEMENTS}
This work was funded in part by NASA Early Career Faculty award 80NSSC20K0070.

\bibliographystyle{IEEEtran}
\balance
\bibliography{references}

\end{document}